\documentclass{article}

\usepackage{arxiv}

\usepackage[utf8]{inputenc} 
\usepackage[T1]{fontenc}    
\usepackage{hyperref}       
\usepackage{url}            
\usepackage{booktabs}       
\usepackage{amsfonts}       
\usepackage{nicefrac}       
\usepackage{microtype}      
\usepackage{lipsum}
\usepackage{graphicx}
\graphicspath{ {./images/} }

\newcommand{\ie}{{\textit{i.e.}}}
\newcommand{\etal}{{\textit{et al.}}}

\usepackage{graphicx}

\usepackage{multicol}
\usepackage{multirow}
\usepackage{amsmath}
\usepackage{makecell}
\usepackage{tabularx}
\usepackage{grffile}
\usepackage[title,toc,titletoc,page]{appendix}
\usepackage[caption=false]{subfig}
\usepackage{pgfplots}
\usepackage{comment}
\usepackage{booktabs}
\usepackage{array}
\newcommand{\PreserveBackslash}[1]{\let\temp=\\#1\let\\=\temp}
\newcolumntype{C}[1]{>{\PreserveBackslash\centering}p{#1}}

\title{Improving Neuroevolution Using Island Extinction and Repopulation\thanks{This material is based upon work supported by the U.S. Department of Energy, Office of Science, Office of Advanced Combustion Systems under Award Number \#FE0031547 and by the Federal Aviation Administration and MITRE Corporation under the National General Aviation Flight Information Database (NGAFID) award.}}

\author{
 Zimeng Lyu \\
  Rochester Institute of Technology\\
  Rochester, NY 14623 \\
  \texttt{zimenglyu@mail.rit.edu} \\
  \And
 Joshua Karns \\
  Rochester Institute of Technology\\
  Rochester, NY 14623 \\
  \texttt{josh@mail.rit.edu} \\
  \And
 AbdElRahman ElSaid \\
  Rochester Institute of Technology\\
  Rochester, NY 14623 \\
  \texttt{aelsaid@mail.rit.edu} \\
  \And
 Travis Desell \\
  Rochester Institute of Technology\\
  Rochester, NY 14623 \\
  \texttt{tjdvse@rit.edu} \\
}

\begin{document}
\maketitle
\begin{abstract}
Neuroevolution commonly uses speciation strategies to better explore the search space of neural network architectures. One such speciation strategy is through the use of islands, which are also popular in improving performance and convergence of distributed evolutionary algorithms. However, in this approach some islands can become stagnant and not find new best solutions. In this paper, we propose utilizing extinction events and island repopulation to avoid premature convergence. We explore this with the Evolutionary eXploration of Augmenting Memory Models (EXAMM) neuro-evolution algorithm. In this strategy, all members of the worst performing island are killed of periodically and repopulated with mutated versions of the global best genome. This island based strategy is additionally compared to NEAT’s (NeuroEvolution of Augmenting Topologies) speciation strategy. Experiments were performed using two different real world time series datasets (coal-fired power plant and aviation flight data). The results show that with statistical significance, this island extinction and repopulation strategy evolves better global best genomes than both EXAMM’s original island based strategy and NEAT’s speciation strategy.

\end{abstract}

\section{Introduction}
    Neuroevolution, or the evolution of artificial neural networks (ANNs) has been widely applied as a neural architecture search strategy for a variety of machine learning problems, including image classification, natural language processing, reinforcement learning and time series data prediction. As the complexity of the tasks ANNs are trained to solve increases, manually designing the network becomes impossible, especially when they may need to be optimized for multiple criteria such as cost, latency, power consumtion and accuracy. Neuroevolution provides a way to evolve ANNs in large and high dimensional space without prior knowledge, searching through the growing number of ANN building blocks, such as activation functions, memory cells, convolutional filter and feature map types, while at the same time determining network topology.

Bio-inspired mata-heuristics are widely used to solve various optimization problems, including the evolution of neural networks, as they are robust enough to solve complicated open-ended questions~\cite{binitha2012survey,darwish2018bio}. ElSaid~\etal~ has utilized ant colony optimization (ACO) to evolve long short-term memory (LSTM) cells for time series data prediction, first evolving LSTM cellular structures~\cite{elsaid2018optimizing} and then later evolving full network topologies~\cite{elsaid2020ant}. Particle swarm optimization (PSO) has also been extended for neuroevolution, where each particle evolves its own network by updating weights and topology~\cite{sahu2016evolving,wang2018evolving,yu2008evolving}. Conforth~\etal~showed that ants and PSO can be combined to evolve ANNs for reinforcement learning tasks~\cite{conforth2008toward}. Other bio-inspired algorithms such as the bird swarm algorithm (BSA)~\cite{aljarah2019evolving} and artificial bee colony (ABC) algorithm~\cite{zhu2019evolutionary} have also been used to evolve ANNs.

Most of the nature inspired EAs start with a random population, and agents in the population evolve network topologies and weights utilizing the bio-inspired rules, retaining the best found solutions. Common to many EA search strategies is the idea of \emph{speciation}, or niching. Within a given evolutionary period, new members of the population are generated by mutation or crossover, have their fitness evaluated, and are inserted into the species which contains the most similar genomes. The mutation, crossover, fitness functions and genome grouping rules are key factors of a speciation strategy, and new speciation strategies can be made by varying those rules. 

As an example, the popular neuroevolution algorithm, NeuroEvolution of Augmenting Topologies (NEAT)~\cite{stanley2002evolving}, seperates genomes into different species by tracking historical genes and measuring the distance between the new genome and an existing species. Another speciation algorithm, Natural Evolution Speciation for NEAT (NENEAT)~\cite{knapp2019natural}, replaces NEAT's speciation with a cladistic strategy where all the genomes in a species share a subset of nodes. Hadjiivanov~\etal~designed a complexity-based speciation strategy, which grouped genomes by the number of hidden neurons~\cite{hadjiivanov2016complexity}. Verbancsics~\etal~investigated the effect of crossover and mutation on neuroevolution speciation strategies~\cite{verbancsics2010evolving}. Sun~\etal~ applied a variable length gene encoding to avoid network depth constraints for solving complex problems~\cite{sun2019evolving}. Krvcah~\etal~modified NEAT's fitness evaluation rule by changing the capacity of species dynamically~\cite{krvcah2012effects}. Instead of using objective functions to measure the fitness of a genome, Lehman~\etal~used searches for behavior novelty~\cite{lehman2011abandoning}. 

Many distributed algorithms utilize the concept of islands, which have been shown by Alba and Tomassini to greatly improve performance of distributed evolutionary algorithms, potentially providing superlinear speedup~\cite{alba2002parallelism}. Islands can also be seen as a speciation strategy, where each island evolves independently and periodically shares genomic information with other islands.

If we look into how species evolve, we find that different species converge and evolve at different speeds. Some species show premature convergence and can become stuck at local optima. Some speciation strategies prevent poorly performing species from reproducing (as in NEAT), however to the authors' knowledge this has not been examined in the context of EAs using a distributed, island-based approach. In this work we take inspiration from \emph{extinction} and \emph{repopulation} events, which have shown to speed up evolution and speciation. Evolving deep neural networks in a large scale is computationally expensive, so if we observe the signs of premature convergence, can we have poorly performing islands go extinct and then be repopulated? Can extinction and repopulation prevent premature convergence and improve the performance of the evolutionary process?

In this paper, we propose a novel repopulation strategy based on extinction events that repopulates poorly performing islands by first removing all the genomes in the island and then repopulating it with random mutations of the global best genome. Experiments explore how the frequency of extinction and the number of random mutations applied to the global best genome affect an island based evolution strategy. This was done using the Evolutionary eXploration of Augmenting Memory Models (EXAMM)~\cite{ororbia2019examm} algorithm that evolves deep Recurrent Neural Networks (RNN) for time series data prediction. We further implemented NEAT's speciation strategy in EXAMM, so it could be fairly compared as a benchmark. To test the robustness of this strategy, we used two real world, non-seasonal, large scale time series data sets from aviation data and a coal-fired power plant. Results show that EXAMM's baseline island based strategy outperforms NEATs strategy with high statistical significance, and further that the new extinction and repopulation based strategies outperform baseline EXAMM, again with statistical significance.

\section{Methodology}
    \subsection{Evolutionary eXploration of Augmenting Memory Models}

This work utilizes the Evolutionary eXploration of Augmenting Memory Models (EXAMM) neuroevolution algorithm to explore extinction and repopulation of islands. EXAMM evolves progressively larger RNNs through a series of mutation and crossover (reproduction) operations. Mutations can be edge-based: \emph{split edge}, \emph{add edge}, \emph{enable edge}, \emph{add recurrent edge}, and \emph{disable edge} operations, or work as higher-level node-based mutations: \emph{disable node}, \emph{enable node}, \emph{add node}, \emph{split node} and \emph{merge node}. The type of node to be added is selected uniformly at random from a suite of simple neurons and complex memory cells: $\Delta$-RNN units~\cite{ororbia2017diff}, gated recurrent units (GRUs)~\cite{chung2014empirical}, long short-term memory cells (LSTMs)~\cite{hochreiter1997long}, minimal gated units (MGUs)~\cite{zhou2016minimal}, and update gate RNN cells (UGRNNs)~\cite{collins2016capacity}. This allows EXAMM to select for the best performing recurrent memory units. EXAMM also allows for \emph{deep recurrent connections} which enables the RNN to directly use information beyond the previous time step. These deep recurrent connections have proven to offer significant improvements in model generalization, even yielding models that outperform state-of-the-art gated architectures~\cite{desell2019evostar-deeprecurrent}.  EXAMM has both a multithreaded implementation and an MPI implementation for distributed use on high performance computing resources. To the authors' knowledge, these capabilities are not available in other neuroevolution frameworks capable of evolving RNNs, which is the primary reason EXAMM was selected to serve as the basis of this work. Due to space limitations we refer the reader to Ororbia \etal~\cite{ororbia2019examm} for more details on EXAMM. 


EXAMM uses an asynchronous island based evolution strategy with a fixed number of islands $n$, each with an island capacity $m$. During the evolution process, islands go through two phases: \emph{initialization}, and \emph{filled}. During the \emph{initialization} phase, each island starts with one seed genome, which is the minimal possible feed-forward neural network structure with no hidden layers, with the input layer fully connected to the output layer. Worker processes repeately request genomes to evaluate from the master process using a work stealing appraoch. 
On receiving a genome the worker then evaluates its \emph{fitness}, calculated as mean squared error (MSE) on a validation data set after stochastic back propagation training. When reported back to the master process, if the island is not full it is inserted into the island, or if the \emph{fitness} is better than the worst genome in that island, it will replace the worst genome. The master generates new genomes from islands in a round-robin manner, by doing $1$ random mutation on randomly selected genomes from an island until that island reaches maximum capacity $m$, and its status becomes \emph{filled}. When all islands are \emph{filled}, they repopulate through inter-island crossover, intra-island crossover and mutation operations. \emph{Intra-island crossover} selects two random genomes from the same island, and the child gets inserted back to where its parents come from. \emph{Inter-island crossover} selects the first parent at random from the target island, and the second parent is the best genome from another randomly selected island. As islands are distinct sub-populations and evolve independently, the only chance for the islands to exchange genes is through \emph{inter-island crossover}. 


The weights of the seed genome generated during the \emph{initialization} phase are initialized uniformly at random between $-0.5$ and $0.5$. After this, RNNs generated through mutation or crossover re-use parental weights, allowing the RNNs to train from where the parents left off, \ie, \emph{“Lamarckian” weight initialization}. Mutation operations add new nodes and edges not present in the parent, and these are initialized using a normal distribution of the average $\mu$ and variance $\sigma^2$ of the best parent's weights. During crossover, in the case where an edge or node exists in both parents, the child weights are generated by recombining the parents’ weights. Given a random number $-0.5<=r<=1.5$, a child’s weight $w_c$ is set to $w_c=r(w_{p2}-w_{p1})+w_{p1}$, where $w_{p1}$ is the weight from the more fit parent, and $w_{p2}$ is the weight from the less fit parent. This allows the child weights to be set along a gradient calculated from the weights of the two parents, allowing for informed exploration of the weight space of the two parents.

\begin{figure*}[ht]
\centering
    \subfloat[\label{fig:example_1} ]{
        \includegraphics[width=0.475\textwidth]{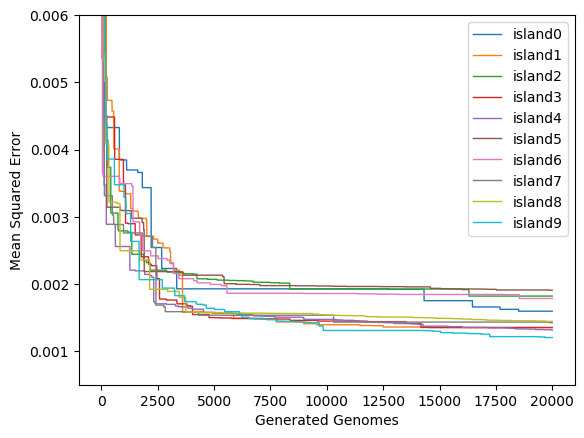}
    }\hfill
    \subfloat[\label{fig:example_2} ]{
        \includegraphics[width=0.475\textwidth]{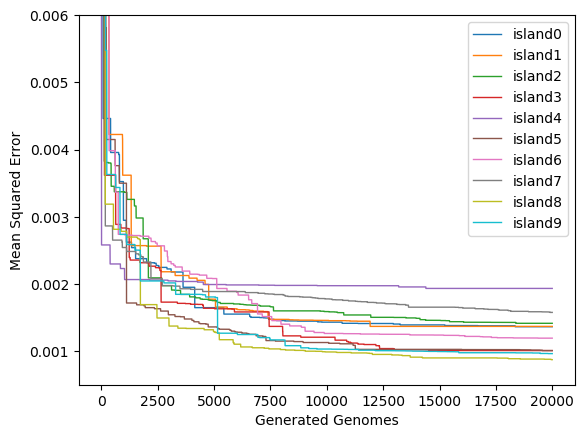}
    }
    \caption{\label{fig:example}Examples of different island convergence rates, with some islands becoming stuck at local optima.}
\end{figure*}

While investigating the performance of the EXAMM algorithm, we observed that islands do not converge at the same speed, and some get stuck. As shown in Figure~\ref{fig:example}, some islands evolve progressively better genomes, while others have premature convergence and get stuck at local optima. As evolving large neural networks is time consuming and computationally expensive, this can waste significant computation resources. If an island cannot reproduce a better genome within certain time period, why not find a way to regenerate or repopulate that island? To accomplish this we examined using \emph{extiction and repopulation} events to help poorly performing islands break out of the local optima. 

\subsection{EXAMM Island Repopulation and Extinction}

A naive approach to repopulation would be to kill off the prematurely converged island and restart from scratch, however, given that the other islands will have well-developed genomes, it might be impossible for it to ever catch up. Further, it would involve re-examining the preliminary regions of the search space. Taking inspiration from nature, most new species are not directly evolved from a single-celled organism. In common cases, a group of organisms evolves in a certain direction to adapt to a new niche, and eventaully a new species emerges. With this as motivation, we utilize the idea of immigrating existing genomes to the worst island for repopulation.  In addition, we examine using mutations on these immigrating genomes to bring innovation through the evolution process, allowing them to poentially further explore new niches.

The EXAMM island repopulation strategy has three phases: \emph{initialization}, \emph{filled}, and \emph{repopulation}. The \emph{initialization} phase is the same as original EXAMM. However, after all the islands become \emph{filled}, we introduce periodic extinction events to the worst performing island. At the time of an extinction event, all the islands are ranked based on their best genome's fitness, and all the genomes in the worst island are removed. Then this island moves into \emph{repopulation} phase. During this phase, new genomes for the island are generated by randomly mutating the global best genome until this island is full and goes back to \emph{filled} status. To handle the asynchronous RNN evaluation in EXAMM, when worker processes return trained RNNs generated from before the extinction event, they are not added to the repopulating island. even With these periodic extinction events, EXAMM encourages further diversity in the entire population.

As a repopulated island might need more time to evolve and find new well performing genomes, if the extinction events keep killing the worst island regardless of if it has just been repopulated, the same island might end up being repeatedly repopulated. On the other hand, mutated global best genomes can perform better or worse than the original one, especially when more than one mutations are applied at one time. If an island has not caught up to the rest of the population by the next extinction event, it may have become stuck in a different local optima. To better examine this, we implemented two repopulation strategies: 1) the worst island can be repopulated at any extinction event, and 2) the worst island can not be repopulated until $5$ extinction events have occurred. In this case, when an extinction event occurs and the island is still the worst, the next worst island which also has not been repopulated within $5$ extinction events will be repopulated.

\subsection{NEAT Speciation}

To provide another benchmark strategy, we also investigated utilizing the speciation strategy from the popular Neuro-Evolution of Augmenting Topologies (NEAT)~\cite{stanley2002evolving}. Instead of using an island strategy, NEAT organizes genomes into small sub-populations, or species. New genomes are inserted into the first species in which the distance $\delta$ between the new genome and a random genome inserted from last generation is less than threshold $\delta_t$.  The distance is calculated using a distance function $\delta$:

    \begin{equation}
        \delta = \frac{C_1E}{N}+\frac{C_2D}{N}+C_3\overline{W}
    \end{equation}
\noindent
where $E$ and $D$ are the excess and disjoint genes between two genomes, and $\overline{W}$ is the weight difference of matching genes. $c1$, $c2$, and $c3$ are hyperparameters which adjust the weight of those factors and $N$ is the number of genes in the larger genome. 

NEAT does not limit the number of species or the species capacity. The species size is controlled by \emph{explicit fitness sharing}~\cite{goldberg1987genetic}. A genome's adjusted fitness ${f'_i}$ is calculated by: 
\begin{equation}
     {f'_i} = \frac{f_i}{\sum_{j=1}^{n}{sh(\delta(i,j))}}
\end{equation}
\noindent
When distance between two genomes $i$ and $j$ exceeds a threshold $\delta_t$, $sh$ is set to 0, $sh$ is 1 otherwise~\cite{spears1995speciation}. Genomes who have a high adjusted fitness ${f'_i}$ are removed. If the best fitness of a species does not improve in 15 generations, this species loses the ability to reproduce. If the entire population does not improve for 20 generations, then only the top 2 species are allowed to reproduce. 



\section{Results}
    \subsection{Data Sets}
This work utilizes two datasets to test the varying speciation strategies. The first comes from a coal-fired power plant (which has requested to remain anonymous) and the second comes from a selection of $10$ flights worth of data from the National General Aviation Flight Information Database (NGAFID). Both datasets are multivariate, with $12$ and $31$ parameters, respectively, non-seasonal, and the parameter recordings are not independent. Furthermore, they are very long – the aviation time series range from $1$ to $3$ hours worth of per-second data while the power plant data consists of $10$ days worth of per-minute readings. \emph{Main flame intensity} was chosen as the prediction parameter from the coal data set, and \emph{pitch} as chosen as the parameter from the flight data set. These data sets are provided openly through the EXAMM GitHub repository\footnote{https://github.com/travisdesell/exact/tree/master/datasets/}.

\subsection{Hyperparameter Settings}
Each EXAMM run used $10$ islands, each with a maximum capacity of $10$ genomes. EXAMM was then allowed to evolve and train $20,000$ genomes through its neuroevolution process. New RNNs were generated via mutation at a rate of 70\%, intra-island crossover at a rate of 20\%, and inter-island crossover at a rate of 10\%. $10$ out of EXAMM's $11$ mutation operations were utilized (all except for \emph{split edge}), and each was chosen with a uniform 10\% chance. EXAMM generated new nodes by selecting from simple neurons, $\Delta$-RNN, GRU, LSTM, MGU, and UGRNN memory cells uniformly at random. Recurrent connections could span any time-skip generated randomly between $\mathcal{U}(1,10)$.

In prior work, EXAMM has been shown to significantly outperform standard NEAT~\cite{elsaid2020ant}, which we attribute mostly to the fact that EXAMM can create nodes from a library of recurrent memory cells, has additional node level mutations, uses a Lamarckian/epigenetic weight inheritance strategy, and trains RNNs via stochastic gradient descent and backpropagation through time (BPTT). On the other hand, NEAT only utilizes edge-level mutations and has a rather simple evolutionary strategy to assign weights to networks.  Additionally, NEAT was not designed for large scale parallelism, and uses a synchronous strategy for iteratively generating new populations. Due to this we implemented NEAT's speciation strategy within the EXAMM framework to compare the speciation strategies without confounding effects from other algorithmic details.

Using recommended hyperparameters, NEAT typically generates $150$ genomes per generation, and if a species has not improved its best fitness within $15$ generations, it will be disabled and not allowed to procreate. It will further disable the entire population except for the top $2$ species if the whole population has not found a new best fitness within $20$ generations. To convert NEAT's generation based strategy to EXAMM's asynchronous strategy, which does not have explicit generations, species were instead disabled if they did not improve after $2250$ new genomes were inserted (the same number of total genomes as $15$ generations of $150$ genomes), and all species except the top $2$ were disabled if the best found fitness did not improve after $3000$ genomes were inserted. The hyperparameters used for NEAT's speciation strategy were $c_1=1$, $c_2=1$, $c_3=0.4$, and the fitness threshold was set to $\delta_t=0.6$ for the coal dataset, and $\delta_t=0.4$ for the flight dataset. The $c_1$, $c_2$ and $c_3$ are standard NEAT values, however the $\delta_t$ values were hand tuned to ensure good speciation. The NEAT runs were highly sensitive to $\delta_t$ and we found higher values resulted in all genomes clustering to the same species, and lower values resulted in each genome having its own species.

For both EXAMM and NEAT, all RNNs were locally trained for $10$ epochs via stochastic gradient descent (SGD) and using back propagation through time (BPTT)~\cite{werbos1990backpropagation} to compute gradients, all using the same hyperparameters. RNN weights were initialized by EXAMM's Lamarckian strategy (described in~\cite{ororbia2019examm}), which allows child RNNs to reuse parental weights, significantly reducing the number of epochs required for the neuroevolution's local RNN training steps. SGD was run with a learning rate of $\eta = 0.001$ and used Nesterov momentum with $\mu = 0.9$. For the memory cells with forget gates, the forget gate bias had a value of $1.0$ added to it (motivated by \cite{jozefowicz2015empirical}).  To prevent exploding gradients, gradient clipping~\cite{pascanu2013difficulty} was used when the norm of the gradient exceeded a threshold of $1.0$. To combat vanishing gradients, gradient boosting (the opposite of clipping) was used when the gradient norm was below $0.05$. These parameters have selected by hand tuning during prior experience with the EXAMM algorithm and these data sets.

\subsection{Experimental Design}

We performed $20$ repeats for each NEAT and EXAMM experiment on the coal and flight data sets. For EXAMM, we compared the baseline strategy (islands without extinction events) to the two variations of the extinction strategy, one allowing repeated repopulations and the other not. For these strategies, extinction frequencies of $1000$ and $2000$ generated genomes were evaluated, and during the repopulation process we allowed the global best genome (at the time of the extinction event) to be mutated either $0$, $2$, $4$, or $8$ times before being inserted into the repopulated island. In total this resulted in a total of $680$ experiments, $20$ for NEAT, $20$ for baseline EXAMM, and $320$ for the $2$ extinction strategies, $2$ extinction frequencies and $4$ mutation values for each of the $2$ datasets.

The different experiments were performed to get an understanding on how the frequency of extinction events effected performance, \ie, did having more frequent extinction events prevent repopulated islands from catching up and improving on the global best solution. Additionally, the two extinction strategies allowed us to determine the impact of allowing islands to be repeatedly made extinct, to see if they needed even more time to become well performing. Finally, modifying the mutation rates was done to provide an idea of how much exploration needed to be performed when repopulating the islands, to allow them to find new potentially better areas of the search space.

\subsection{Computing Environment}
Results were gathered using Rochester Institute of Technology's research computing systems. This system consists of 2304 Intel® Xeon® Gold 6150 CPU 2.70GHz cores and 24 TB RAM, with compute nodes running the RedHat Enterprise Linux 7 system. All EXAMM baseline and EXAMM speciation strategies experiments utilized 180 cores. Since the NEAT speciation strategy is implemented in the EXAMM framework, and the EXAMM master process is responsible for generating and inserting genomes, whereas worker processes are only responsible for stochastic back propagation training and evaluate the fitness of genomes, all the genome distances and explicate fitness sharing evaluations were done in the master process. Utilizing NEAT's speciation strategy presented a speed bottleneck at the master process when using a larger number of cores. Due to this we used 72 cores for all the NEAT runs, as adding additional cores did not improve runtime.

\subsection{Repopulation Strategy Evaluation}

\begin{figure*}[ht]
\centering
    \subfloat{
        \includegraphics[width=0.475\textwidth]{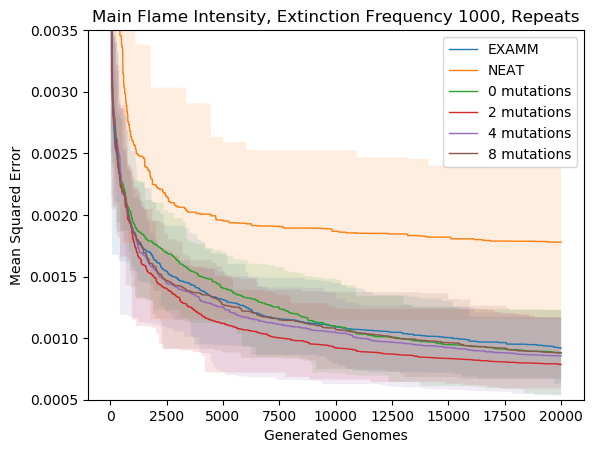}
    }\hfill
    \subfloat{
        \includegraphics[width=0.475\textwidth]{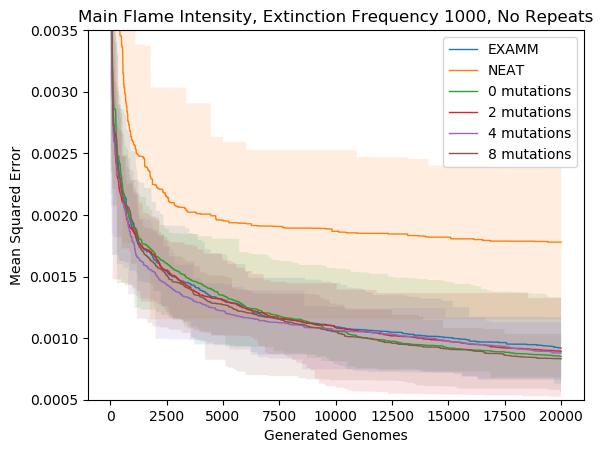}
    }

    \subfloat{
        \includegraphics[width=0.475\textwidth]{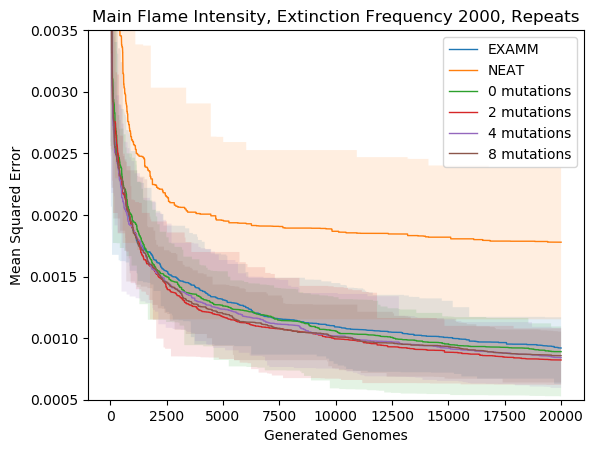}
    }\hfill
    \subfloat{
        \includegraphics[width=0.475\textwidth]{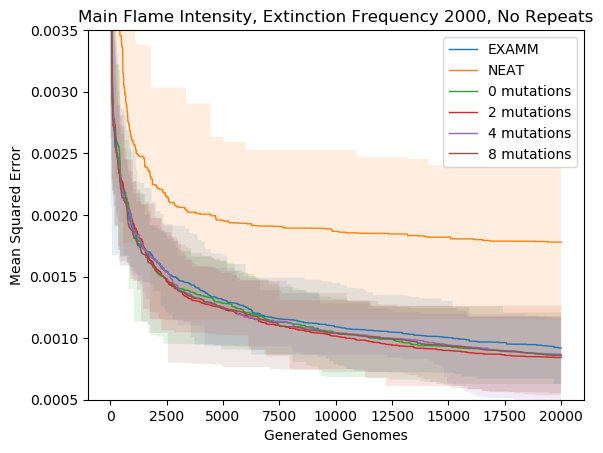}
    }

    \caption{\label{fig:coal} Convergence rates (in terms of best MSE on validation data) for NEAT speciation and the EXAMM extinction and repopulation strategies predicting \emph{main flame intensity} from the  coal fired power plant dataset.}
\end{figure*}

\begin{figure*}[ht]
\centering
    \subfloat{
        \includegraphics[width=0.475\textwidth]{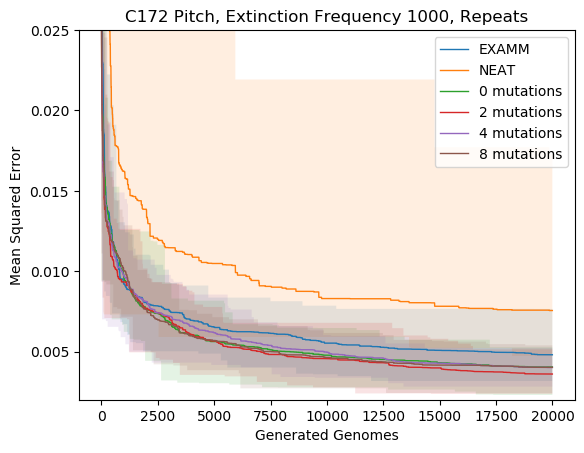}
    }\hfill
    \subfloat{
        \includegraphics[width=0.475\textwidth]{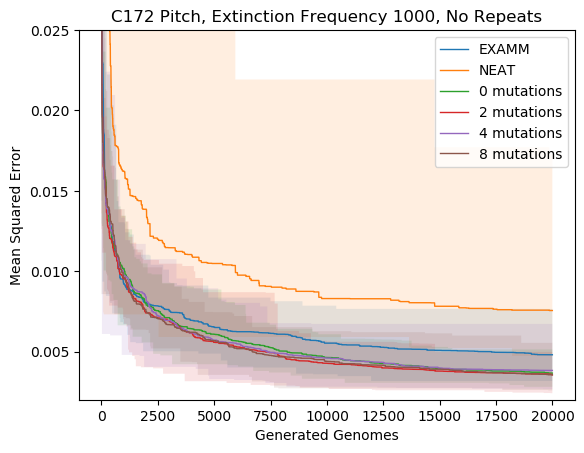}
    }

    \subfloat{
        \includegraphics[width=0.475\textwidth]{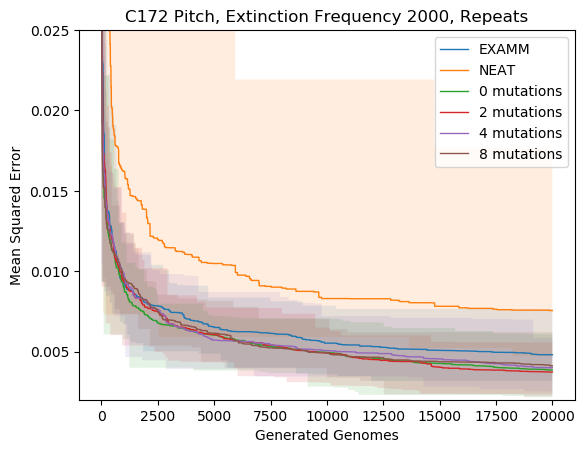}
    }\hfill
    \subfloat{
        \includegraphics[width=0.475\textwidth]{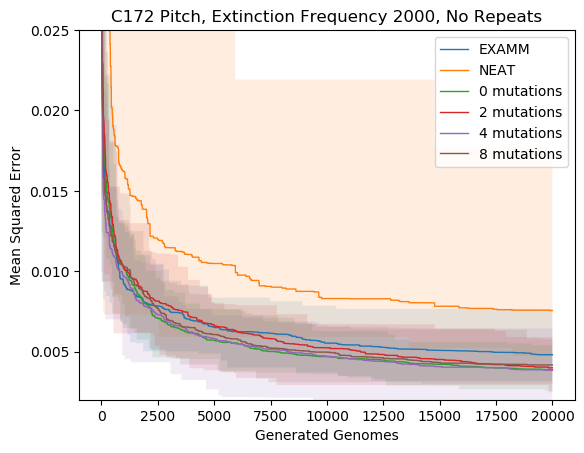}
    }

    \caption{\label{fig:flight}Convergence rates (in terms of best MSE on validation data) for NEAT speciation and the EXAMM extinction and repopulation strategies predicting \emph{pitch} from the c172 flight dataset.}
\end{figure*}

Figures~\ref{fig:coal} and~\ref{fig:flight} present results showing the performance across the $10$ repeated experiments for NEAT speciation and the EXAMM variations. The solid line shows the average of the global best genomes across the $10$ experiments, and the filled in area shows the range between the min and max. The test results show that the EXAMM extinction and repopulation strategies perform better than baseline EXAMM algorithm across all tests, with the NEAT speciation strategy performing worse than baseline EXAMM. On average, in three of the coal plant test cases and two of the four flight test cases, $2$ mutations finds the best performing genomes. For the other test cases, applying $4$ or $8$ mutations finds the best performing genomes, and all those test cases come from non-repeated repopulations for both datasets, which proves that innovations need more time to evolve and become better. The results also suggest that adding some, but not too much variance to the global best genome for island repopulation allowed the strategies to best find new regions of the search space to improve performance.

\begin{table}
\centering
    \begin{tabular}{|C{40pt}|c|c|C{50pt}|C{40pt}|C{40pt}|C{40pt}|C{40pt}|} \hline
            &   Erase         &    Extinct &  &  & & &\\
     Dataset &  Rule & Freq  & Mutation & p-Value & Worst  & Avg  & Best\\ \hline \hline

        \multirow{18}{*}{Coal} & EXAMM & / &/ & / & 0.00117 & 0.00092 & 0.00063 \\ \cline{2-8}  
                                &    NEAT & / & / & {\bf3.9e-8} & 0.00238 & 0.00178 & 0.00115\\ \cline{2-8}  
                                                         
                          & \multirow{8}{*}{Repeat} & \multirow{4}{*}{1000} & 0 & 0.2124 & 0.00123 & 0.00088 & 0.00054\\ \cline{4-8}
                          &                         &                       & 2 & {\bf0.0014} & 0.00117 & {\bf 0.00079} & 0.00059\\ \cline{4-8}
                          &                         &                       & 4 & 0.1427 & 0.00116 & 0.00086 & 0.00054\\ \cline{4-8}
                          &                         &                       & 8 & 0.1897 & 0.00123 & 0.00088 & 0.00067\\ \cline{3-8}
                          &                         & \multirow{4}{*}{2000} & 0 & 0.3180 & 0.00109 & 0.00089 & 0.00053\\ \cline{4-8}
                          &                         &                       & 2 & {\bf0.0266} & 0.00108 & 0.00082 & 0.00062\\ \cline{4-8}
                          &                         &                       & 4 & {\bf0.0283} & 0.00108 & 0.00084 & 0.00060\\ \cline{4-8}
                          &                         &                       & 8 & 0.1092 & 0.00105 & 0.00086 & 0.00064\\ \cline{2-8}

            &\multirow{8}{*}{\shortstack{No\\Repeat}}& \multirow{4}{*}{1000} & 0 & {\bf0.0360} & 0.00133 & 0.00085 & 0.00068\\ \cline{4-8}
                          &                         &                       & 2 & 0.3474 & 0.00133 & 0.00089 & 0.00052\\ \cline{4-8}
                          &                         &                       & 4 & 0.1308 & 0.00113 & 0.00088 & 0.00066\\ \cline{4-8}
                          &                         &                       & 8 & {\bf0.0405} & {\bf0.00103} & 0.00083 & 0.00057\\ \cline{3-8}
                          &                         & \multirow{4}{*}{2000} & 0 & 0.0702 & 0.00121 & 0.00087 & 0.00063\\ \cline{4-8}
                          &                         &                       & 2 & 0.0568 & 0.00126 & 0.00084 & 0.00054\\ \cline{4-8}
                          &                         &                       & 4 & 0.1197 & 0.00116 & 0.00087 & {\bf0.00049}\\ \cline{4-8}
                          &                         &                       & 8 & 0.0632 & 0.00118 & 0.00086 & 0.00055\\ \hline
        \multirow{18}{*}{C172} &  EXAMM & / &/ & / & 0.00765 & 0.00480 & 0.00316 \\ \cline{2-8}
                          &     NEAT & / & / & {\bf1.3e-6}  & 0.01725 & 0.00755 & 0.00473\\ \cline{2-8}  
                          & \multirow{8}{*}{Repeat} & \multirow{4}{*}{1000} & 0 & {\bf0.0057} & 0.00526 & 0.00404 & 0.00229\\ \cline{4-8}
                          &                         &                       & 2 & {\bf3.3e-5} & 0.00514 & 0.00360 & 0.00236\\ \cline{4-8}
                          &                         &                       & 4 & {\bf0.0026} & 0.00523 & 0.00401 & 0.00282\\ \cline{4-8}
                          &                         &                       & 8 & {\bf0.0057} & 0.00538 & 0.00401 & 0.00242\\ \cline{3-8}
                          &                         & \multirow{4}{*}{2000} & 0 & {\bf0.0022} & 0.00606 & 0.00385 & 0.00223\\ \cline{4-8}
                          &                         &                       & 2 & {\bf0.0002} & 0.00556 & 0.00371 & 0.00234\\ \cline{4-8}
                          &                         &                       & 4 & {\bf0.0128} & 0.00584 & 0.00399 & 0.00252\\ \cline{4-8}
                          &                         &                       & 8 & {\bf0.0104} & 0.00621 & 0.00411 & 0.00216\\ \cline{2-8}
                         &\multirow{8}{*}{\shortstack{No\\Repeat}}& \multirow{4}{*}{1000} & 0 & {\bf3.7e-5} & 0.00513 & 0.00366 & 0.00281\\ \cline{4-8}
                          &                         &                       & 2 & {\bf3.7e-5}  & {\bf 0.00497} & {\bf 0.00355} & 0.00240\\ \cline{4-8}
                          &                         &                       & 4 & {\bf4.2e-4}  & 0.00672 & 0.00382 & 0.00266\\ \cline{4-8}
                          &                         &                       & 8 & {\bf4.6e-5} & 0.00554 & {\bf 0.00355} & 0.00258\\ \cline{3-8}
                          &                         & \multirow{4}{*}{2000} & 0 & {\bf0.0005}  & 0.00539 & 0.00387 & 0.00257\\ \cline{4-8}
                          &                         &                       & 2 & {\bf0.0030} & 0.00590 & 0.00398 & 0.00246\\ \cline{4-8}
                          &                         &                       & 4 & {\bf0.0011} & 0.00643 & 0.00381 & {\bf 0.00163}\\ \cline{4-8}
                          &                         &                       & 8 & {\bf0.0049} & 0.00577 & 0.00411 & 0.00300\\ 
     
        \hline
    \end{tabular}
    \caption{\label{table:p_values} 
    Performance of the various strategies, including worst, average and best-case validation mean squared error on the flight dataset for the varying EXAMM experiments, with best performance for this dataset marked bold. Mann–Whitney U test $p$-values are included comparing EXAMM to NEAT speciation and the different extinction and repopulation strategies. $p$-values in bold indicate a statistically significant difference with $\alpha = 0.05$.}
\end{table}

As a further investigation, Table~\ref{table:p_values} presents Mann–Whitney U-test $p$-values comparing the best genomes of the $20$ repeats of varying strategies to the best genomes from the $20$ repeats of baseline EXAMM. $p$-values in bold represent statistically significant differences with $\alpha = 0.05$, showing that the results of the varying mutation strategies have a statistically significant difference from EXAMM, which similarly has a statistically significant difference from NEAT speciation.

Table~\ref{table:p_values} also provides more detail about the best, average and worst global best genome fitness at the end of the 20 repeated tests for each experiment. From this we can see that in the average cases having a faster extinction frequency of $1000$ generally provided the best results, providing more evidence that performing extinction and repopulation is improving the performance of neuroevolution strategy. Interestingly, the strategy which allowed islands to not be repeatedly killed provided slightly better results in the best case for both the coal and flight data. Also it is worth noting that, having 4 mutations and an extinction frequency of $2000$ did find the best genome for both datasets, suggesting that having more mutations can potentially stumble upon better search areas.



\section{Conclusion}

This work investigates a novel speciation strategy based on extinction and repopulation events for island based evolutionary algorithms, applying it to neuroevolution of recurrent neural networks for time series data prediction on two challenging real-world data sets. In this strategy, the worst performing islands are periodically killed off and repopulated with either the global best genome or mutations of it. Two versions of this were implemented, one which allowed islands to be repeatedly repopulated and the other which prevented an island being repopulated until a specified number of extinction events occurred on other islands. We investigated versions of this strategy with varying extinction frequencies, as well as numbers of mutations to the global best genome.

These mutation strategies were incorporated into the Evolutionary eXploration of Augmenting Memory Models (EXAMM) neuroevolution project, along with NEAT's speciation strategy as a benchmark comparison to a well-known neuroevolution technique. Results show that that the repopulation strategy led to statistically significant improvements over baseline EXAMM, which in turn had large and statistically significant improvements over NEAT's speciation strategy. It was found that the number of mutations applied to the global genome during repopulation were not significantly correlated with the best performance, however in general a lower number (2 or 4) provided the best results. Having more mutations brought more innovation, but was also more unstable, leading to repopulated islands being repeatedly killed. Allowing islands to be repeatedly repopulated had advantages and disadvantages, where repeated repopulation would remove ``bad" genomes more quickly, but preventing repeat repopulations protected innovations, giving the repopulated islands more time to evolve. In general, both strategies (allowing and disallowing repeated repopulation) provided statistically significant improvements over not using extinction and repopulation, but interestingly, neither significantly outperformed the other.

This work explored how different number of mutations combined with different island extinction rules effected the repopulation process. Future work will involve examining other types of island extinction events, for example killing off multiple islands during an extinction, or controlling extinction events based on how much an island has improved over a period of time. Other options for repopulation can also be investigated beyond using the global best genome. Future work will also includes investigating how to use varying forms of crossover to improve the repopulation algorithm's performance, which will include examining various crossover rules for repopulation, changing genome encoding methods, and redesigning the distance evaluation function. Lastly, it was particularly interesting that preventing and allowing repeated extinction both provided similar improvements but neither outperformed the other. Developing a strategy which can make use of the best qualities of both may lead to further performance improvements. It should also be noted that while this work was examind in the context of neuroevolution algorithms, it could also be applied to any evolutionary strategy utilizing islands.


\section*{Acknowledgements}

Most of the computation of this research was done on the high performance computing clusters of Research Computing at Rochester Institute of Technology~\cite{https://doi.org/10.34788/0s3g-qd15}. We would like to thank the Research Computing team for their assistance and the support they generously offered to ensure that the heavy computation this study required was available.

\bibliographystyle{unsrt}  

\newpage

\bibliography{./references.bib} 

\end{document}